\documentclass[letterpaper, 10 pt, conference]{ieeeconf}  %

\IEEEoverridecommandlockouts                              %

\usepackage{amsmath,amssymb,amsfonts}
\usepackage{graphicx}
\usepackage{hyperref}
\usepackage[caption=false, font=footnotesize]{subfig}

\title{Deep 2.5D Vehicle Classification with Sparse SfM Depth Prior for Automated Toll Systems\\
}

\author{Georg Waltner$^{1}$, Michael Maurer$^{1}$, Thomas Holzmann$^{1}$, Patrick Ruprecht$^{1}$, Michael Opitz$^{1}$,\\ Horst Possegger$^{1}$, Friedrich Fraundorfer$^{1}$ and Horst Bischof$^{1}$%
\thanks{This work was supported by the Austrian Research Promotion Agency (FFG) under the project \textit{3DMaut (856161)}. We gratefully acknowledge the support of NVIDIA Corporation with the donation of the Titan X Pascal GPU used for this research.}%
\thanks{$^{1}$All authors are with Institute of Computer Graphics and Vision, Faculty of Computer Science, Graz University of Technology, 8010 Graz, Austria
        {\tt\small waltner@icg.tugraz.at}}%
}

\begin{document}

\maketitle
\thispagestyle{empty}
\pagestyle{empty}

\begin{abstract}

Automated toll systems rely on proper classification of the passing vehicles. This is especially difficult when the images used for classification only cover parts of the vehicle. To obtain information about the whole vehicle. we reconstruct the vehicle as 3D object and exploit this additional information within a Convolutional Neural Network (CNN). However, when using deep networks for 3D object classification, large amounts of dense 3D models are required for good accuracy, which are often neither available nor feasible to process due to memory requirements. Therefore, in our method we reproject the 3D object onto the image plane using the reconstructed points, lines or both. We utilize this sparse depth prior within an auxiliary network branch that acts as a regularizer during training. We show that this auxiliary regularizer helps to improve accuracy compared to 2D classification on a real-world dataset. Furthermore due to the design of the network, at test time only the 2D camera images are required for classification which enables the usage in portable computer vision systems.

\end{abstract}

\section{INTRODUCTION}

\begin{figure*}[htpb]
	\centerline{\includegraphics[width=.75\textwidth]{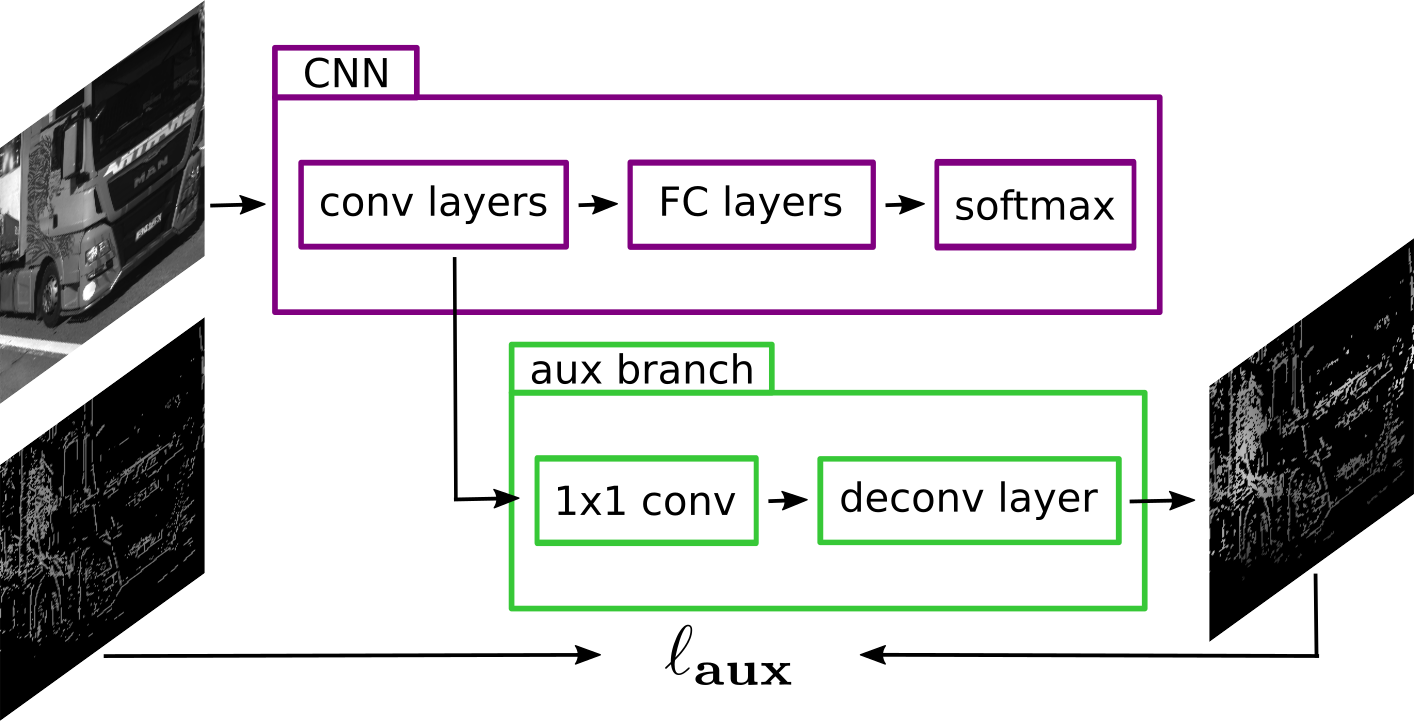}}
	\caption{Modified CNN network for classification with SfM prior. We add an auxiliary branch to incorporate the 2.5D depth information into the CNN. After the last convolutional layer, we add a $1\times1$ convolution and an deconvolution layer. The loss $\ell_{\mathbf{aux}}$ is then calculated between the upsampled layer activations and the groundtruth generated from our 3D reconstruction. We use a VGG-16 model~\cite{rel-vgg16}, but any classification CNN can be used.}
	\label{fig:vgg-model}
\end{figure*}
In recent years, electronic systems replaced manual toll collection to eliminate delays and traffic congestions on toll roads. At first, self-service toll booths were installed, where the driver pays the toll with cash or credit card. In this scenario the driver still needs to stop and process the payment at a machine, thus not resolving delay and congestion problem. After introducing automated toll collection systems, the vehicles are only required to slow down to a certain speed and pass the toll section. 
The collection is either done via a pay-by-plate system where a computer vision system recognizes the license plate for billing or a transponder system where the billing is initiated after passing a gating system. These gating systems are built along the toll road and the vehicle passes them without further impact on the driving behavior.
Along with the automation of toll collection it has become more difficult to control the payments made to the operator, as the human as controlling factor has been mostly removed from the toll collecting system. An electronic toll device might deliberately report a false vehicle type to the toll system and pay less fees.
Therefore, different measures try to prevent fraud or inaccurate reporting of toll obligations.\\
One of them is computer vision aided control, where a camera system checks whether the reported toll information is correct or not. With some imagery at hand, a classification algorithm decides which category a seen vehicle belongs to. 
Designing such computer vision systems is a challenging task. To train complex classifiers like a Convolutional Neural Network (CNN), thousands of images need to be collected and labeled. Since typically only parts of the vehicle are visible, the available 2D information might be insufficient for classification due to ambiguities.
To overcome these problems, a 3D reconstruction of objects can be used for classification. Such reconstructions can be obtained by applying Structure-from-Motion (SfM) on a sequence of many images. In real world applications like ours, often only very few images are available. This results in reconstructions of 3D models that are very limited in completeness and density, even after additional post-processing. While there has been progress with 3D convolutional neural networks recently, classifiers operating on such sparse 3D models often do not perform well enough in terms of accuracy to be employed in real-world applications. In addition, required hardware resources are mostly not available on site.\\
To overcome these limitations, our approach utilizes 2D and 3D information in an efficient way. We propose to use the sparse depth data as auxiliary loss to improve the classification accuracy of a CNN. Therefore we obtain a sparse point cloud from a SfM pipeline and project these 3D points into the camera views. We also use 3D lines for the projections to capture vehicle structure.
This yields a 2.5D representation that we feed as a sparse depth prior along with the recorded images into a CNN for classification. See Fig.~\ref{fig:vgg-model} for an overview of our CNN model structure.\\
The main benefit of our method is that we are able to efficiently leverage the 3D vehicle structure information in addition to the 2D appearance information. As we show in our experiments, by using depth as auxiliary loss we can significantly improve the accuracy of a CNN. Further, since we do not need the depth map during test time, we do not have to run a computationally expensive SfM pipeline. Consequently, our approach can run on embedded hardware in a portable toll control system.

\section{RELATED WORK}
We focus our summary of related work on the different topics our method relates to. First, we introduce the used SfM methods, then we give a short overview of 2D and 3D classification algorithms based on CNNs. %
\subsection{Structure-from-Motion (SfM)}
Structure-from-Motion is a technique to reconstruct a 3D model from 2D images. In most cases, thousands of images are required to output a good representation of the object. A typical workflow consists of the following steps: First, keypoints are detected at image locations that are distinguishable by their gradients (e.g. corners). The regions around those keypoints are described with SIFT~\cite{rel-sift} or SURF~\cite{rel-surf} features such that each point is represented by a vector of same length and thus comparable with a distance metric. Matching keypoints between pairs of images are then found based on the feature vector distance. %
From these matches, the five-point algorithm~\cite{rel-5point} estimates and verifies the relative motion between image pairs.
In a final optimization step called bundle-adjustment~\cite{rel-bundle-adjustment}, the camera poses are refined such that the triangulation error of the 3D points is minimized. The final result is a 3D point cloud, where every point can be seen from at least two images of the dataset. In this work we use the algorithm of~\cite{rel-irschara} to obtain the sparse point cloud with oriented camera poses.\\
In our work, complementary to point clouds we also use 3D reconstructions consisting of lines. We use the method of~\cite{rel-l3d++}, where 2D line segments are detected and then matched in 3D using geometric constraints. These constraints are defined by the camera poses generated from the SfM pipeline.
\begin{figure*} 
	\centering
	\subfloat[Image 1]{%
		\includegraphics[width=0.19\linewidth]{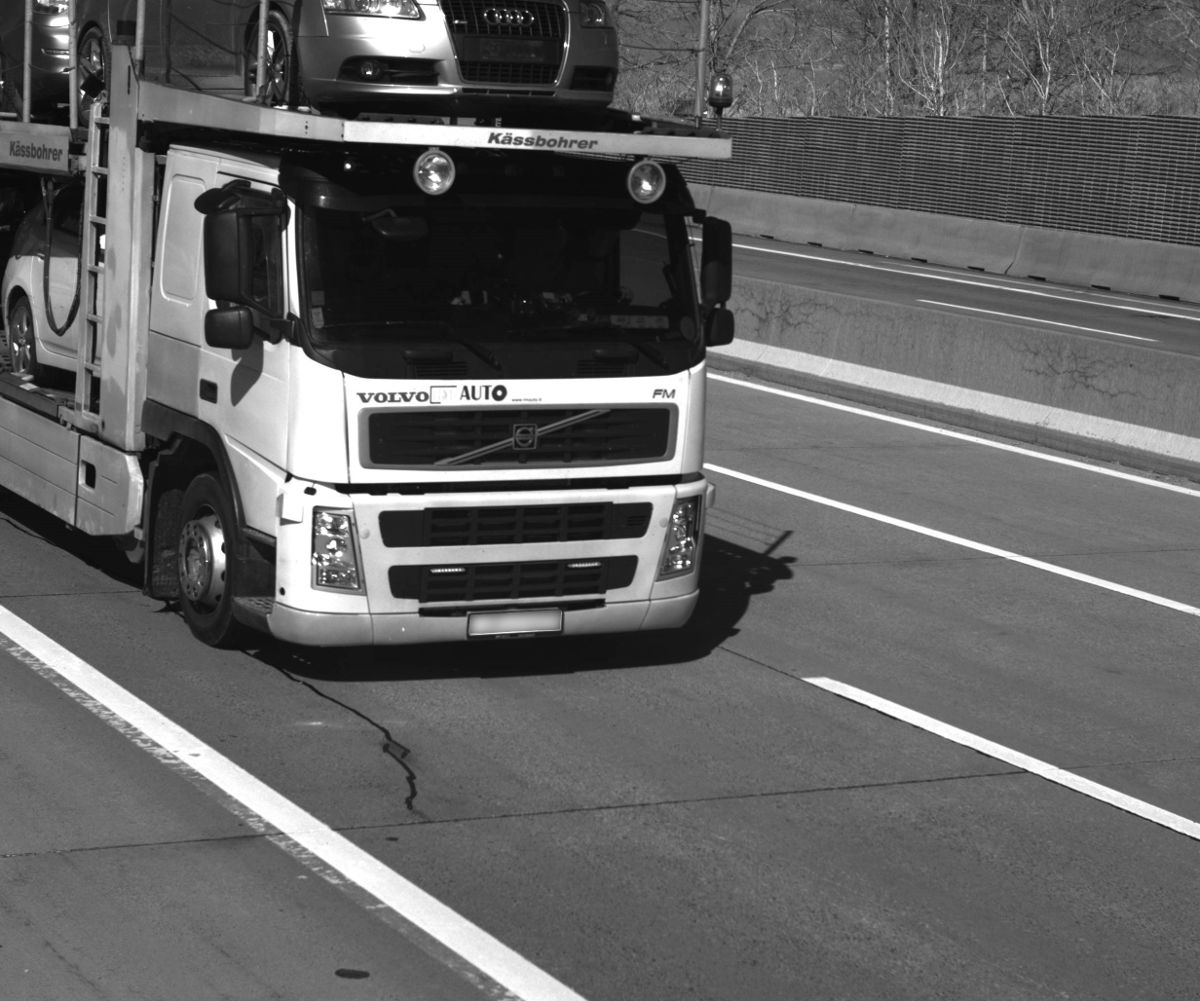}}
	\hfill
	\subfloat[Image 2]{%
		\includegraphics[width=0.19\linewidth]{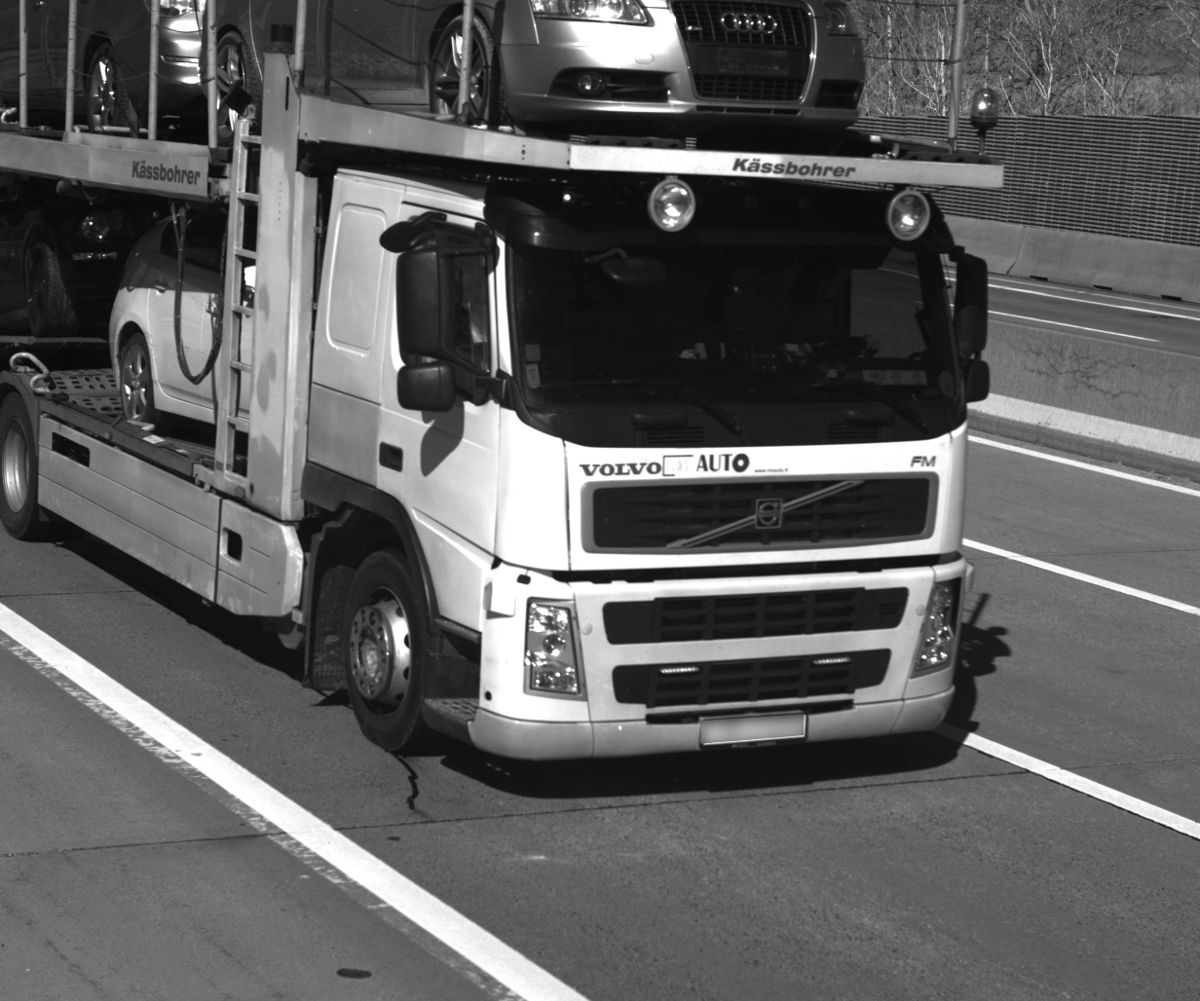}}
	\hfill
	\subfloat[Image 3]{%
		\includegraphics[width=0.19\linewidth]{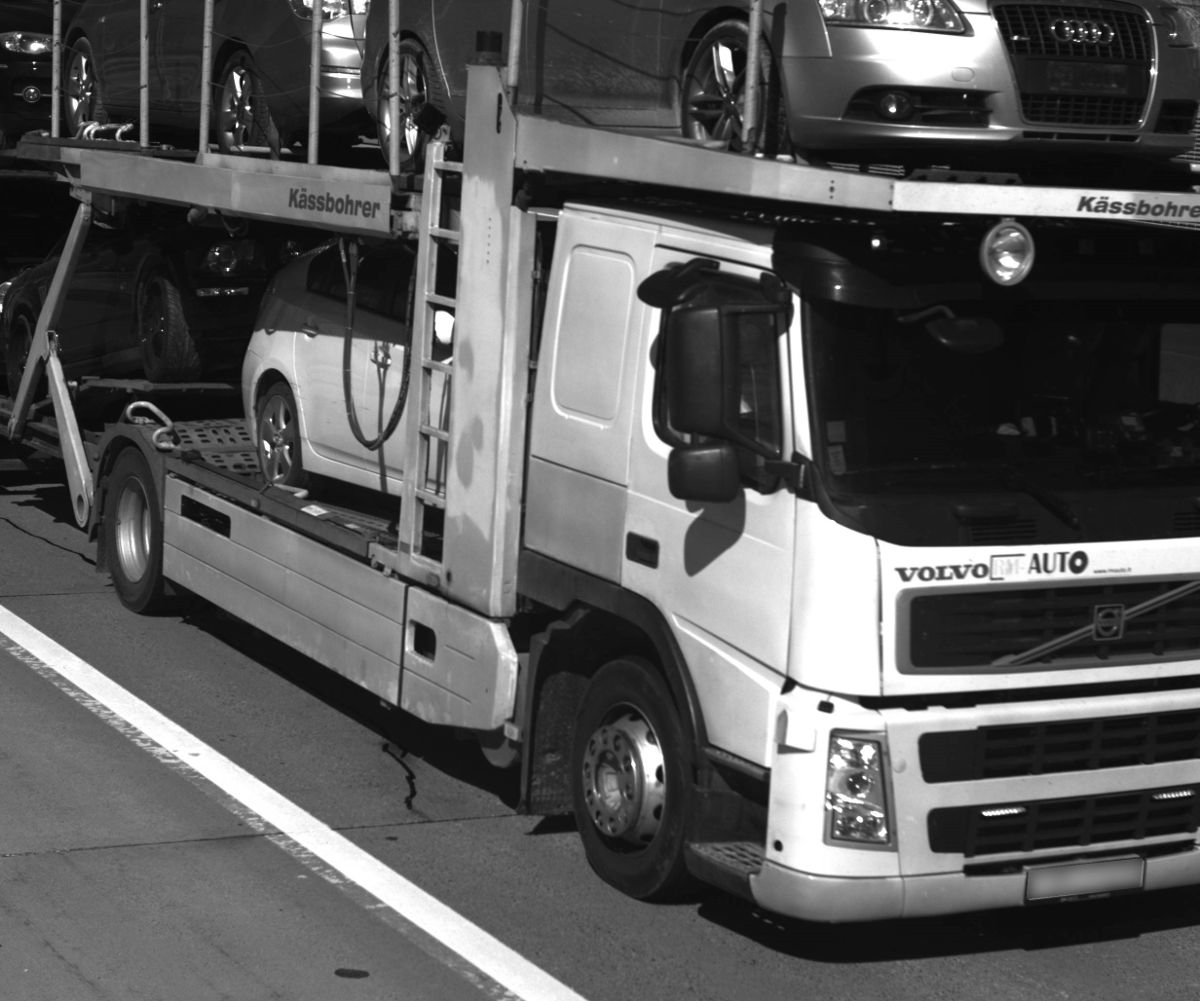}}
	\hfill
	\subfloat[Image 4]{%
		\includegraphics[width=0.19\linewidth]{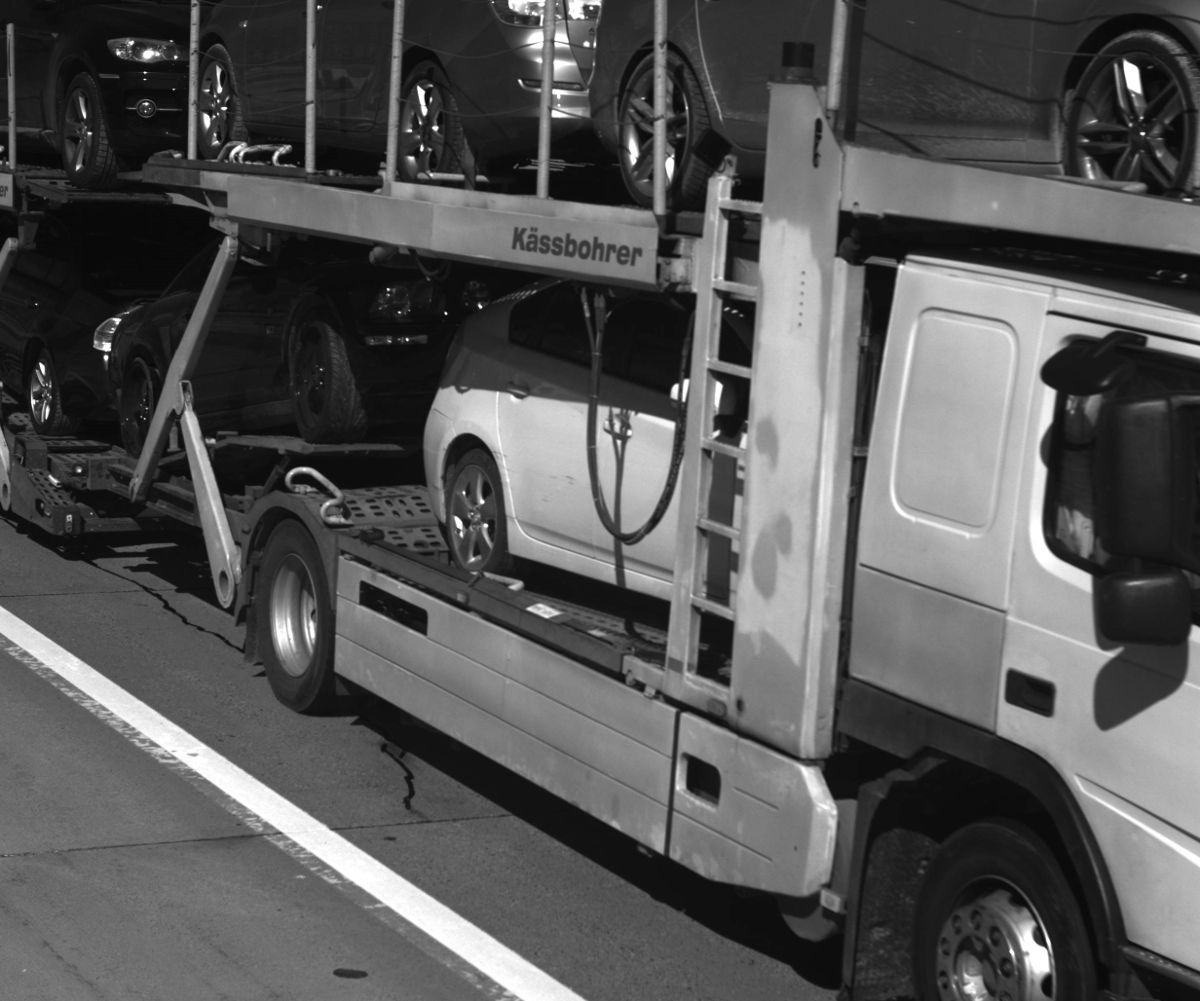}}
	\hfill
	\subfloat[Image 5]{%
		\includegraphics[width=0.19\linewidth]{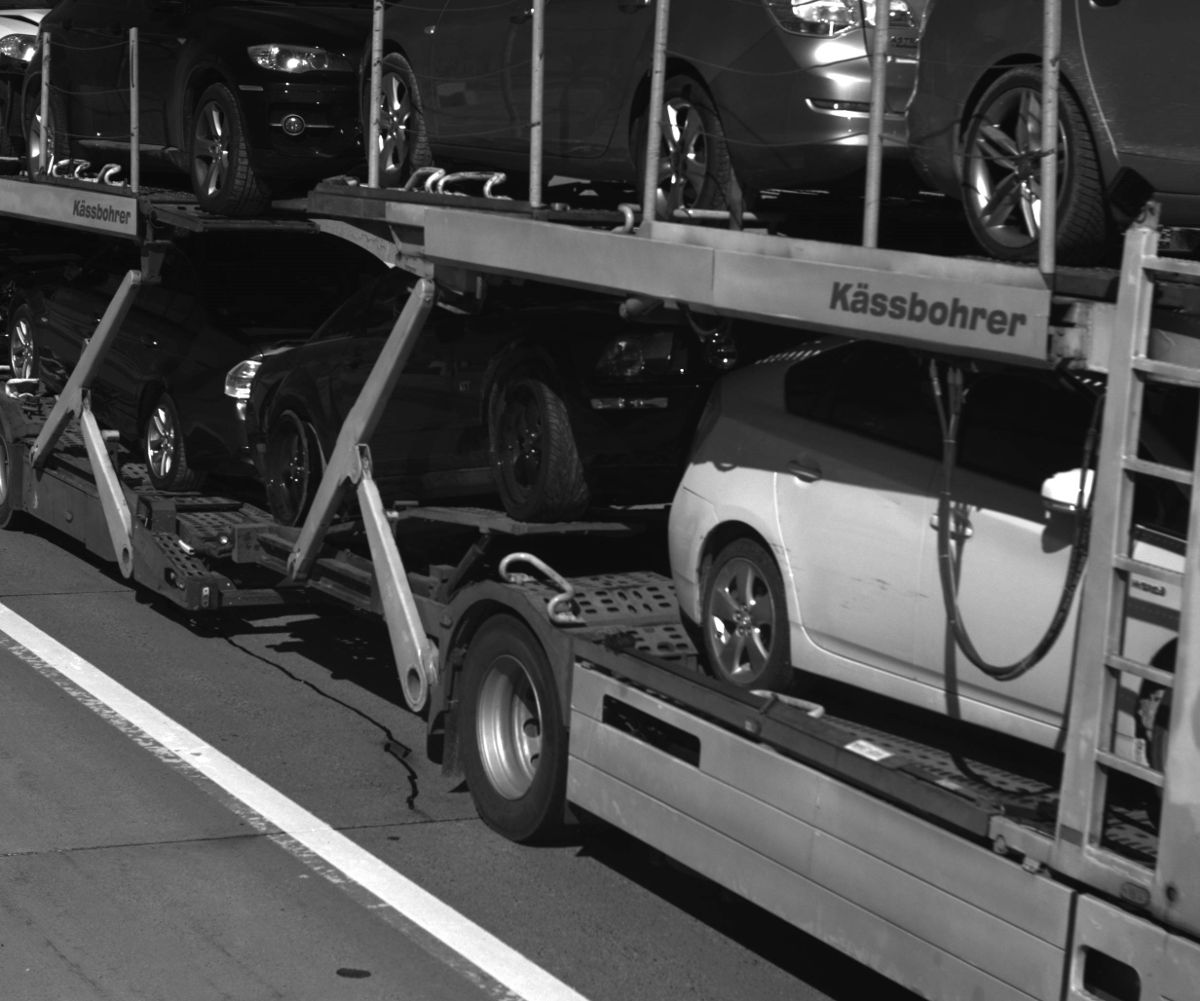}}
	\hfill\\
	\subfloat[Image 6]{%
		\includegraphics[width=0.19\linewidth]{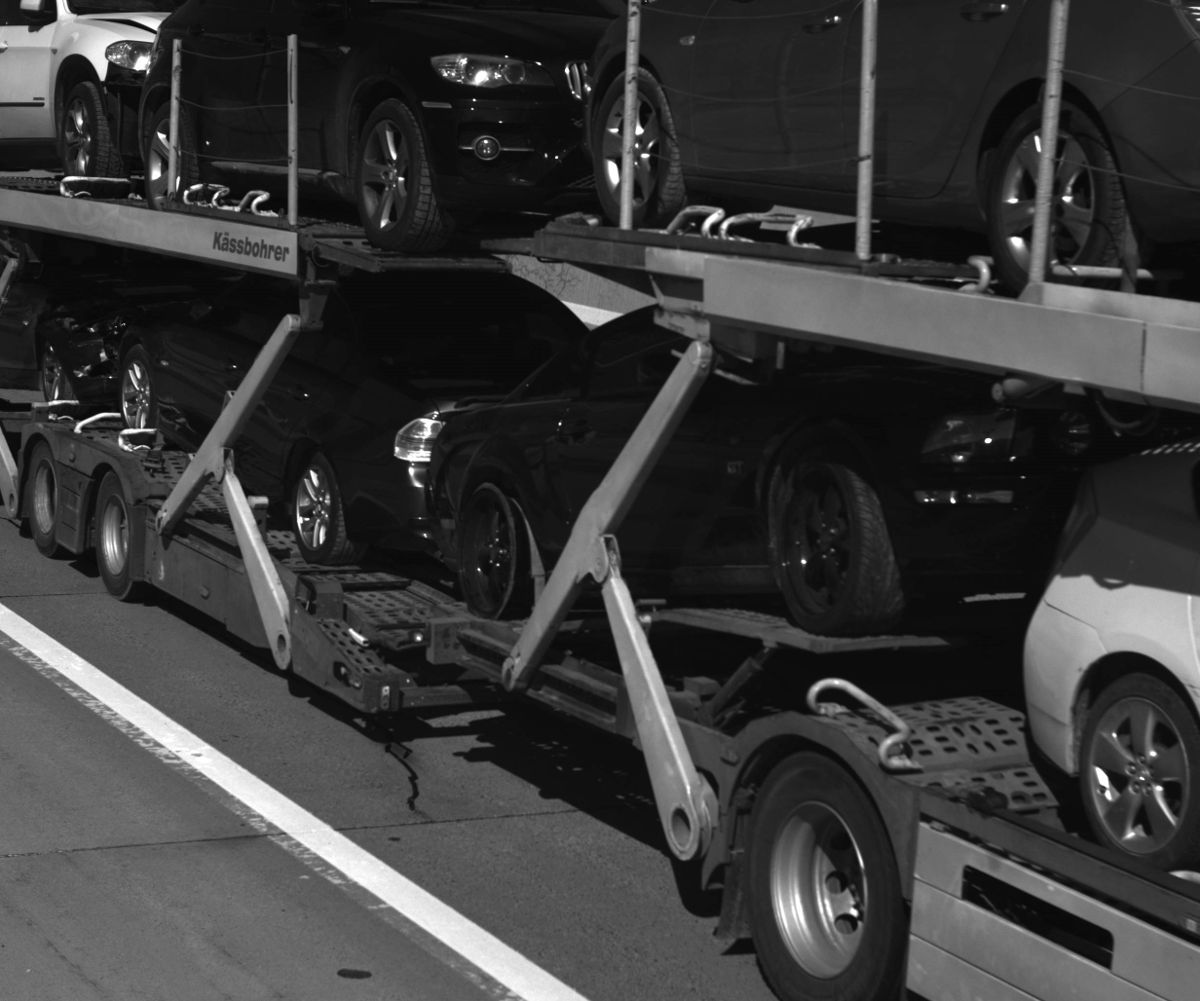}}
	\hfill
	\subfloat[Image 7]{%
		\includegraphics[width=0.19\linewidth]{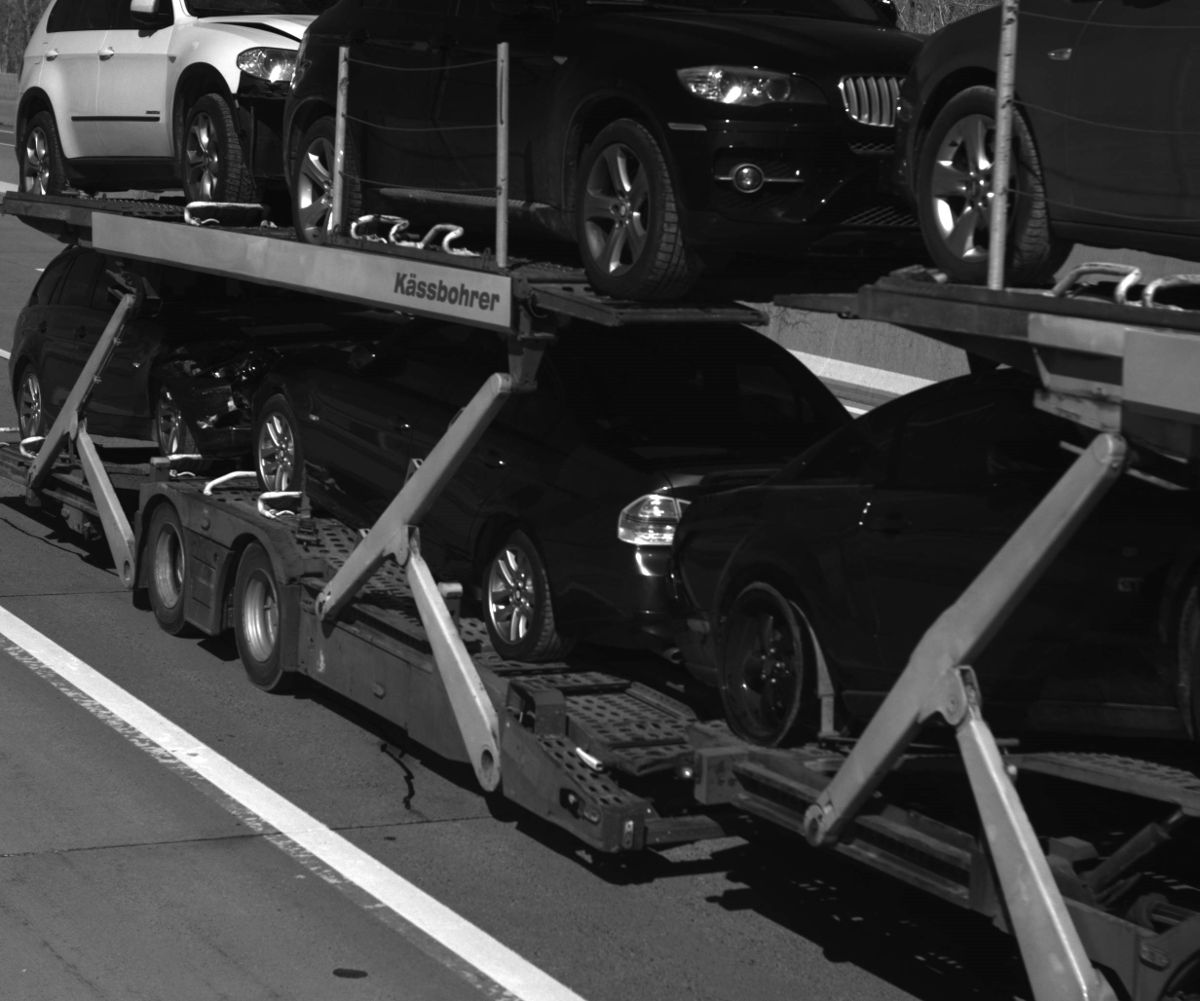}}
	\hfill
	\subfloat[Image 8]{%
		\includegraphics[width=0.19\linewidth]{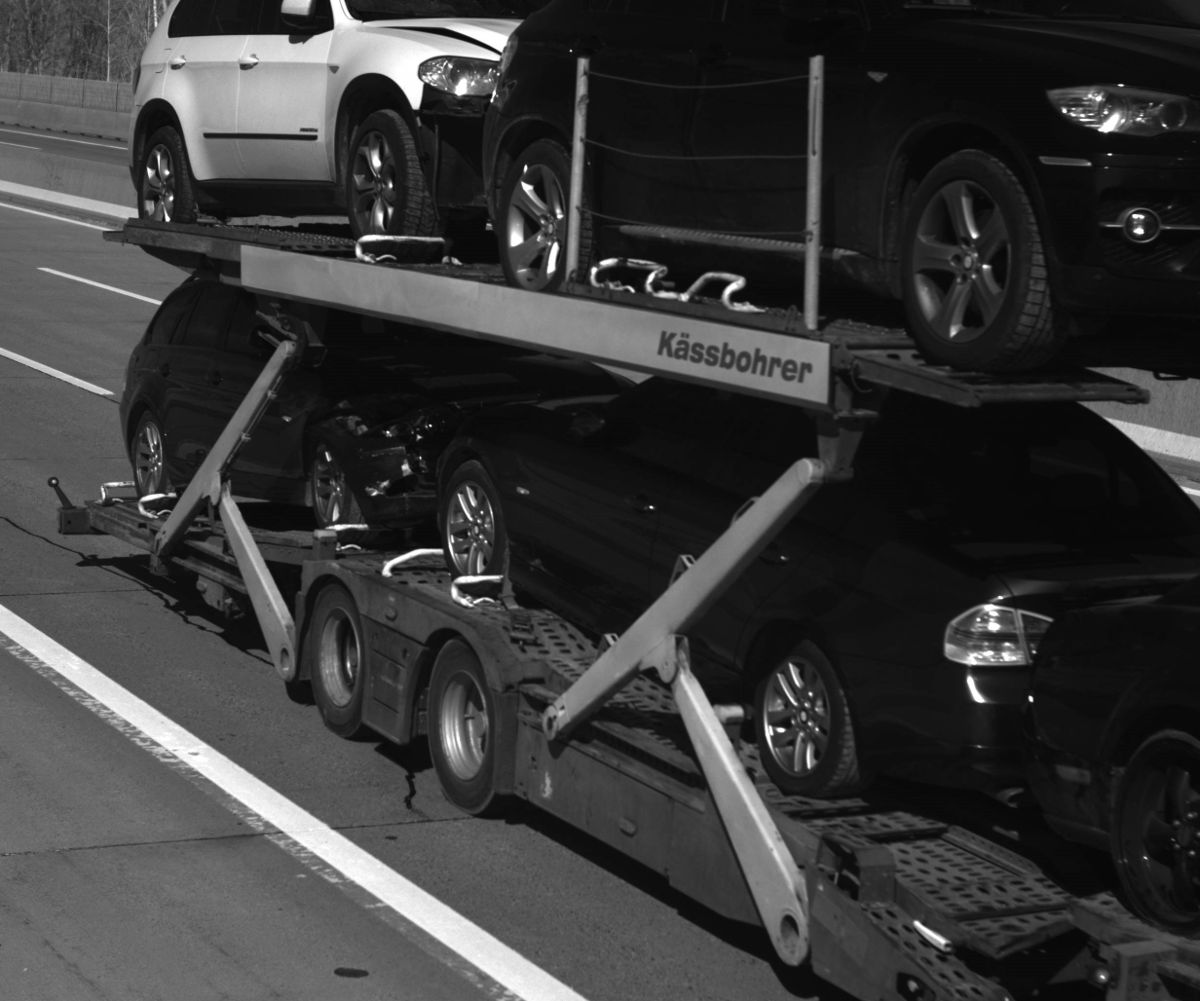}}
	\hfill
	\subfloat[Image 9]{%
		\includegraphics[width=0.19\linewidth]{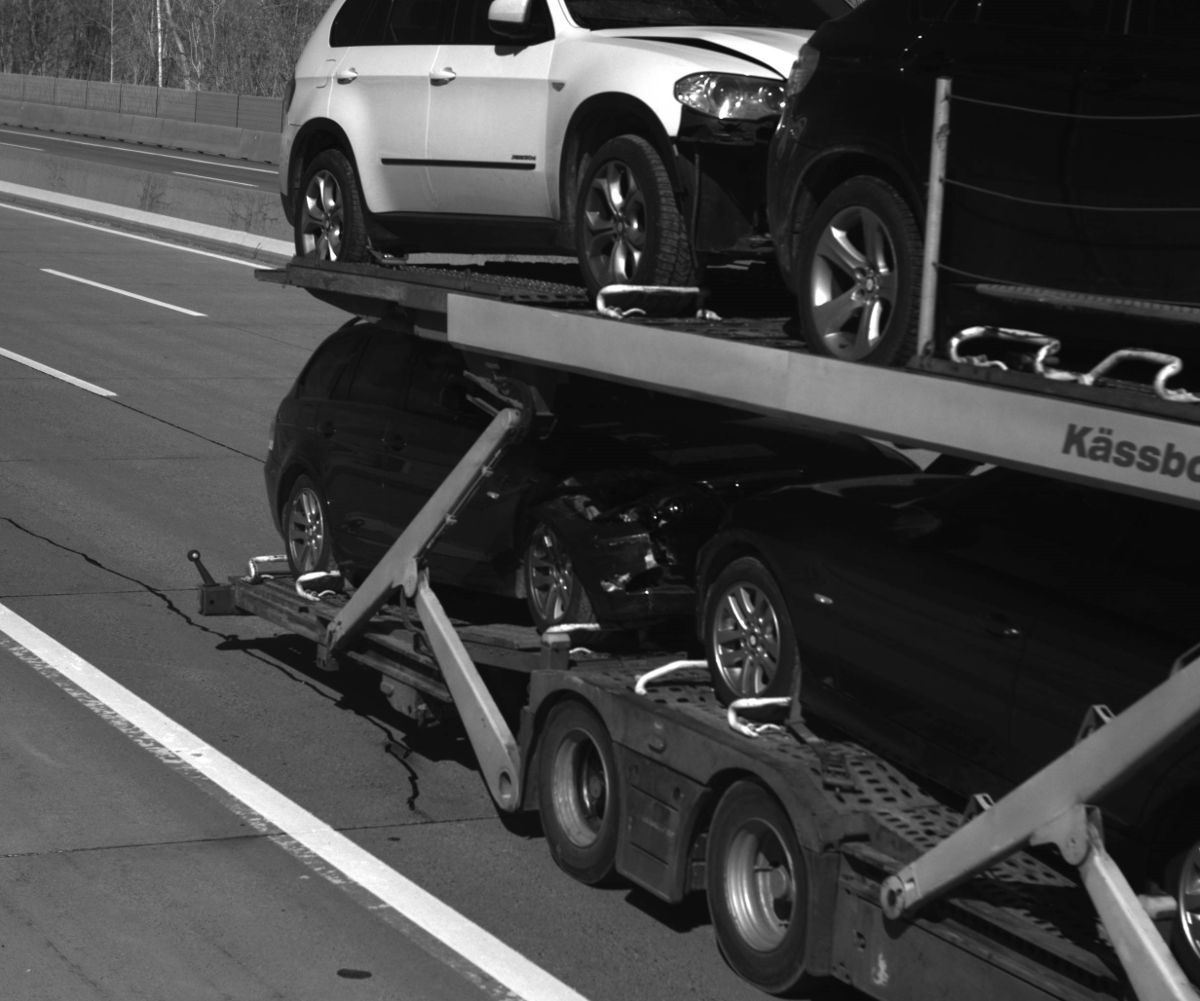}}
	\hfill
	\subfloat[Image 10]{%
		\includegraphics[width=0.19\linewidth]{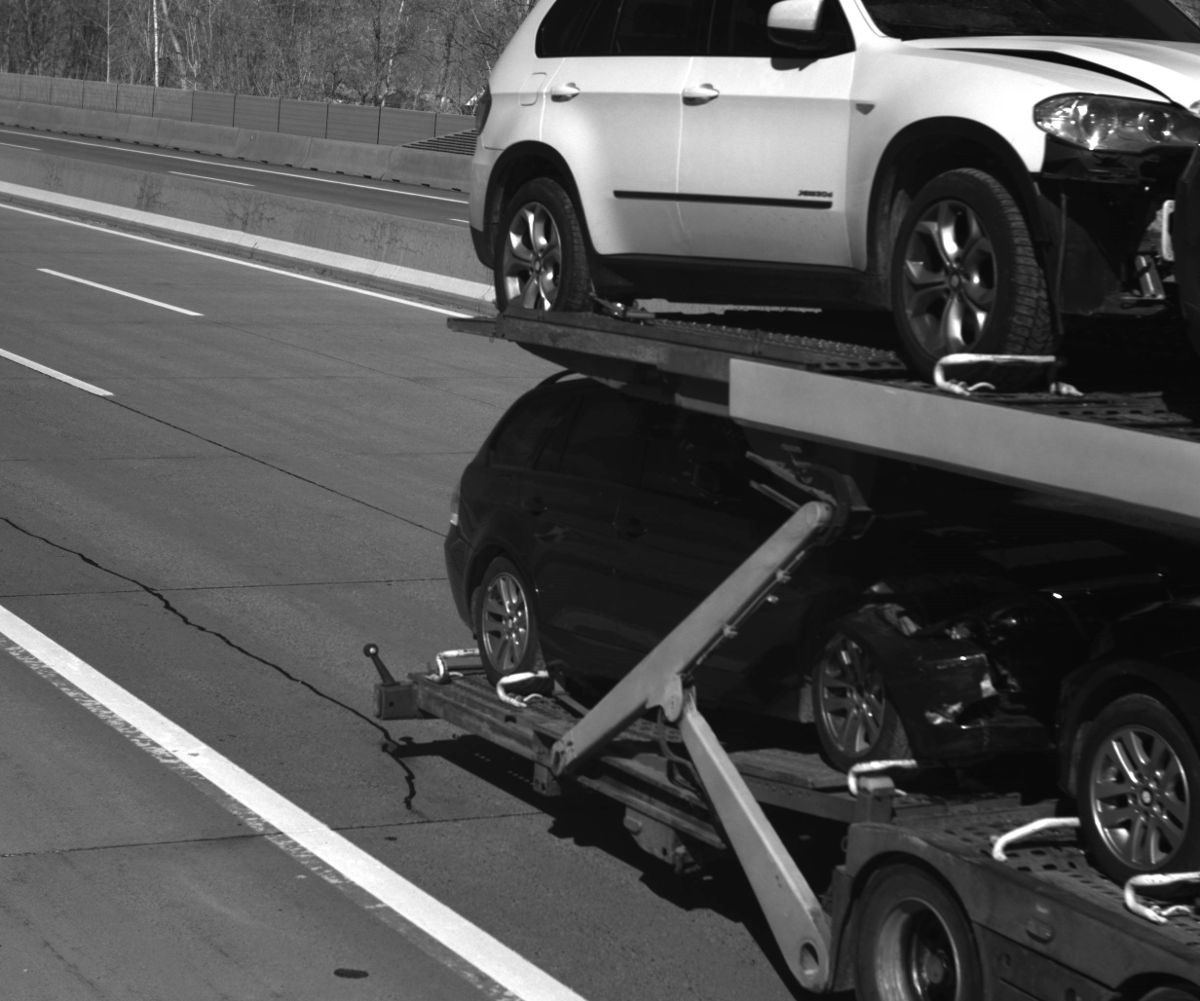}}
	\hfill\\
	\subfloat[Angled view]{%
		\includegraphics[width=0.38\linewidth]{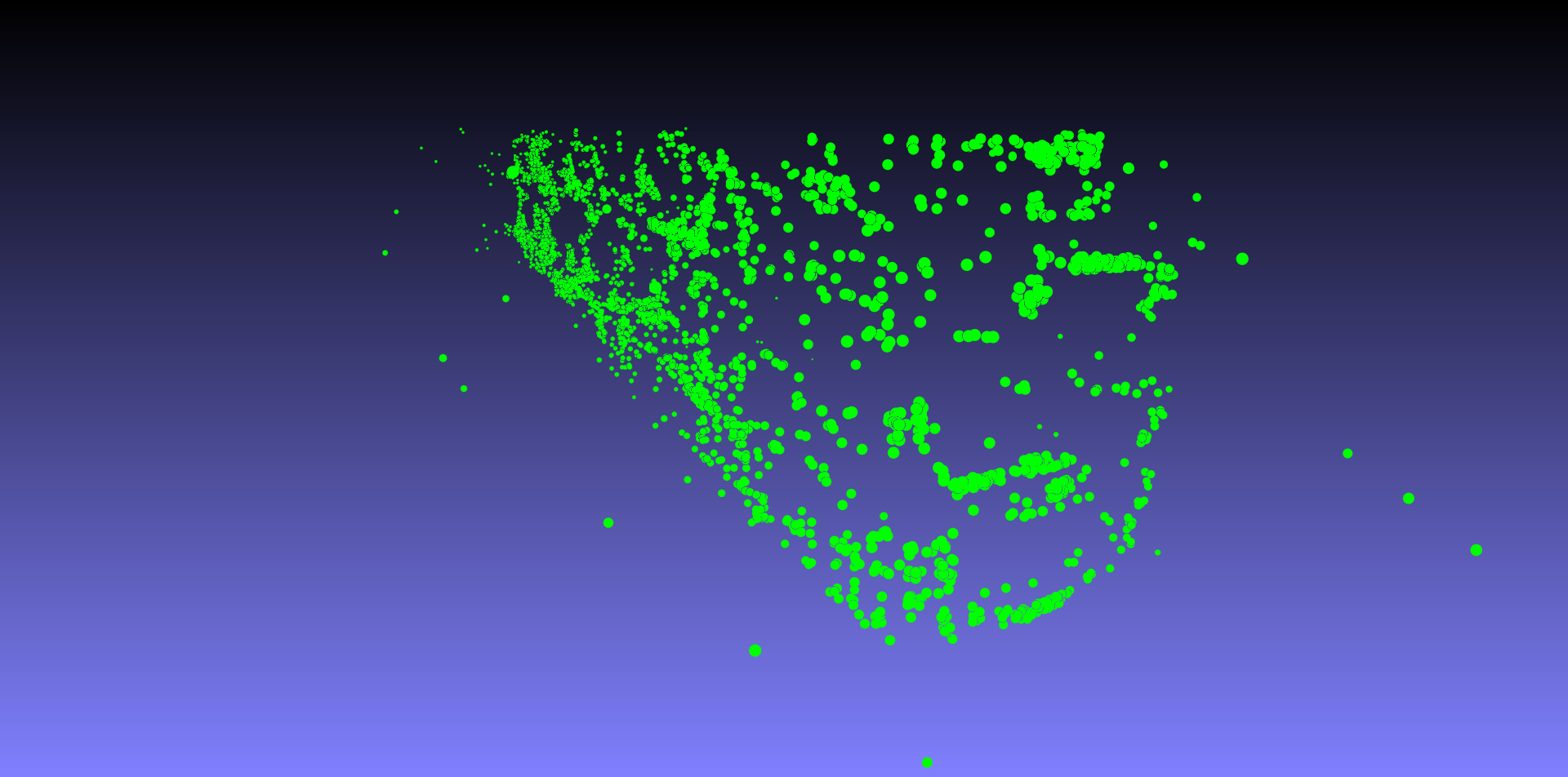}}
	\hspace{40pt}
	\subfloat[Side view]{%
		\includegraphics[width=0.38\linewidth]{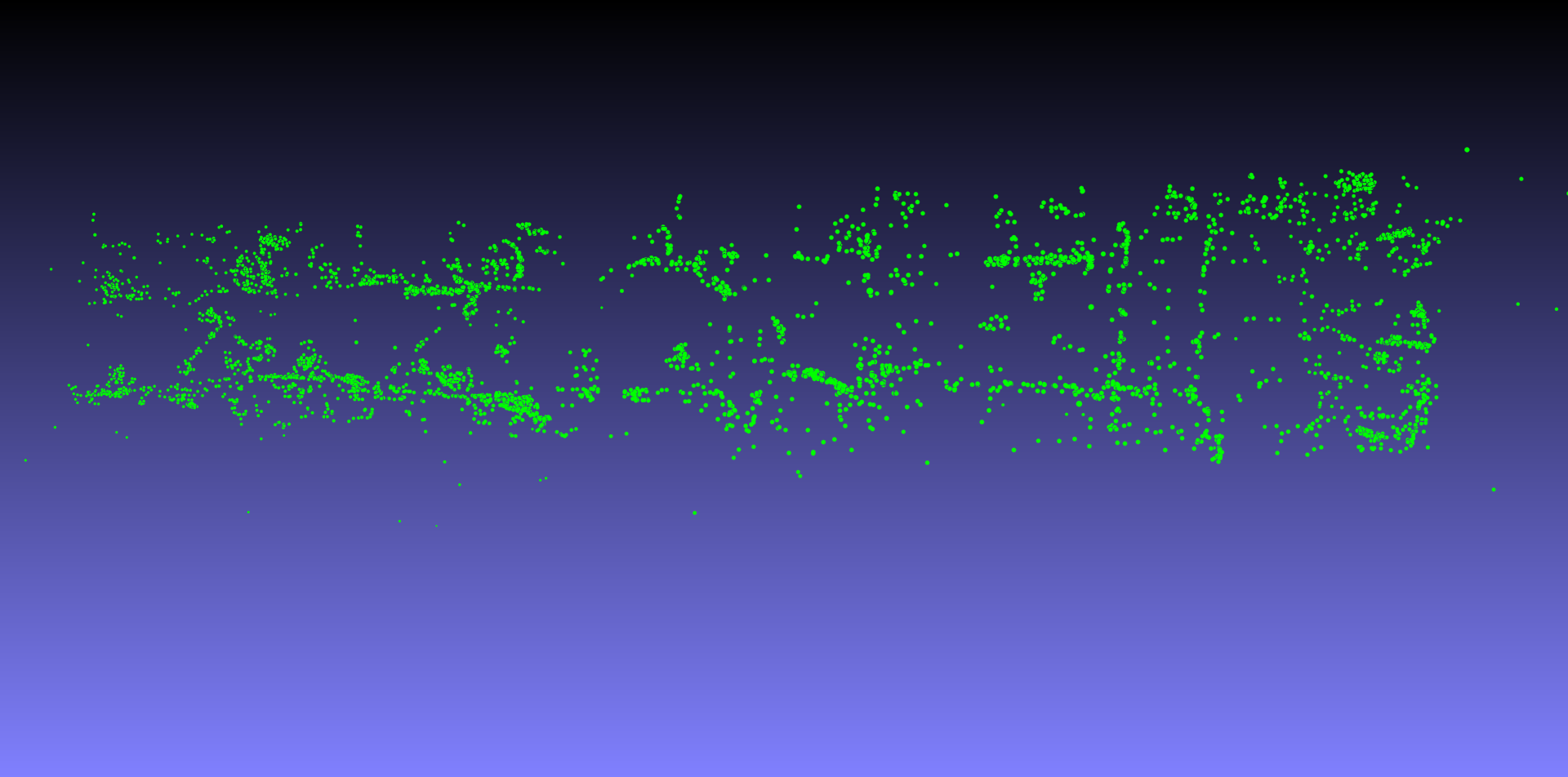}}
	\hfill
	\subfloat[Top view]{%
		\includegraphics[width=0.38\linewidth]{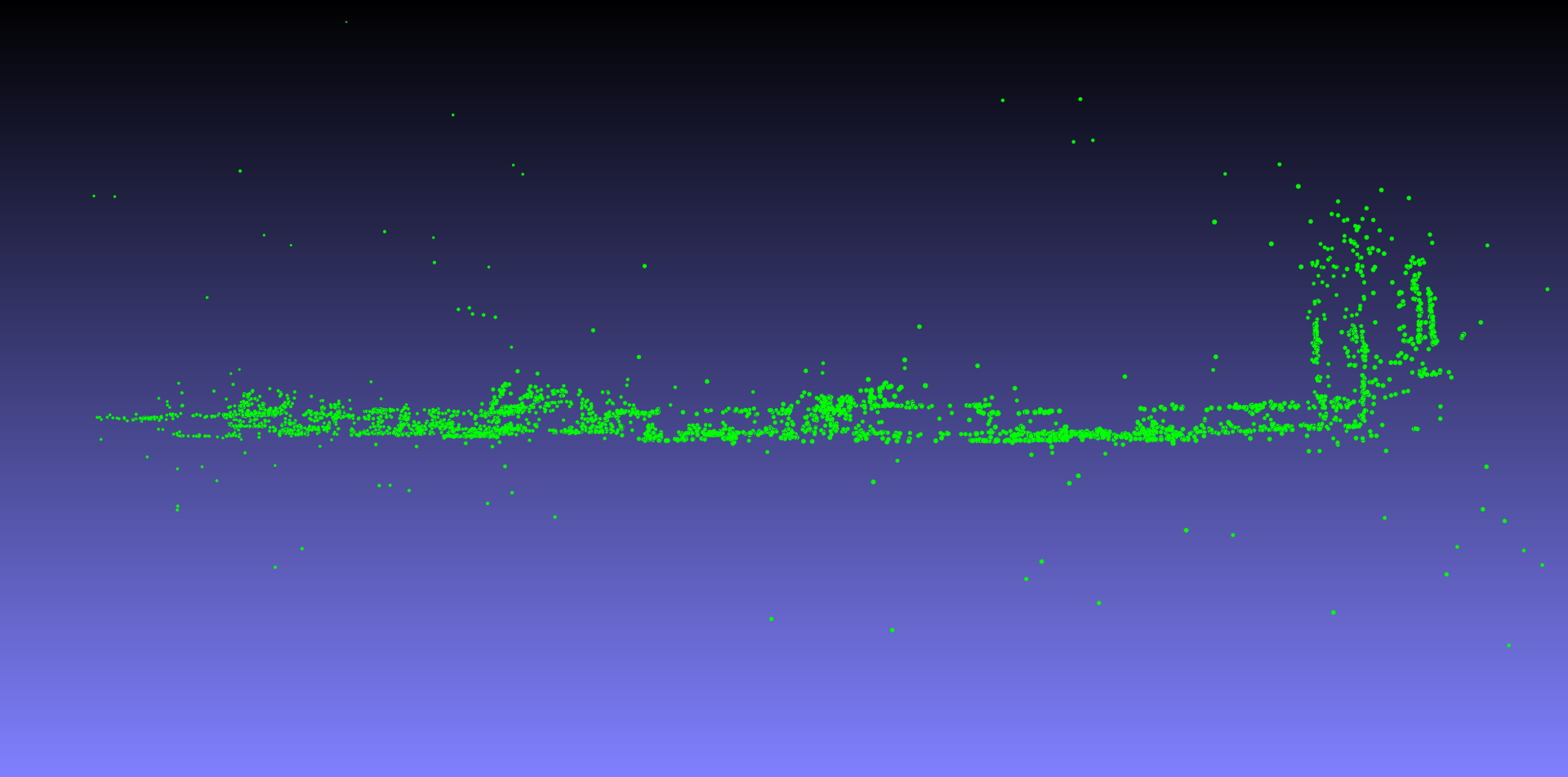}}
	\hspace{40pt}
	\subfloat[Front view]{%
		\includegraphics[width=0.38\linewidth]{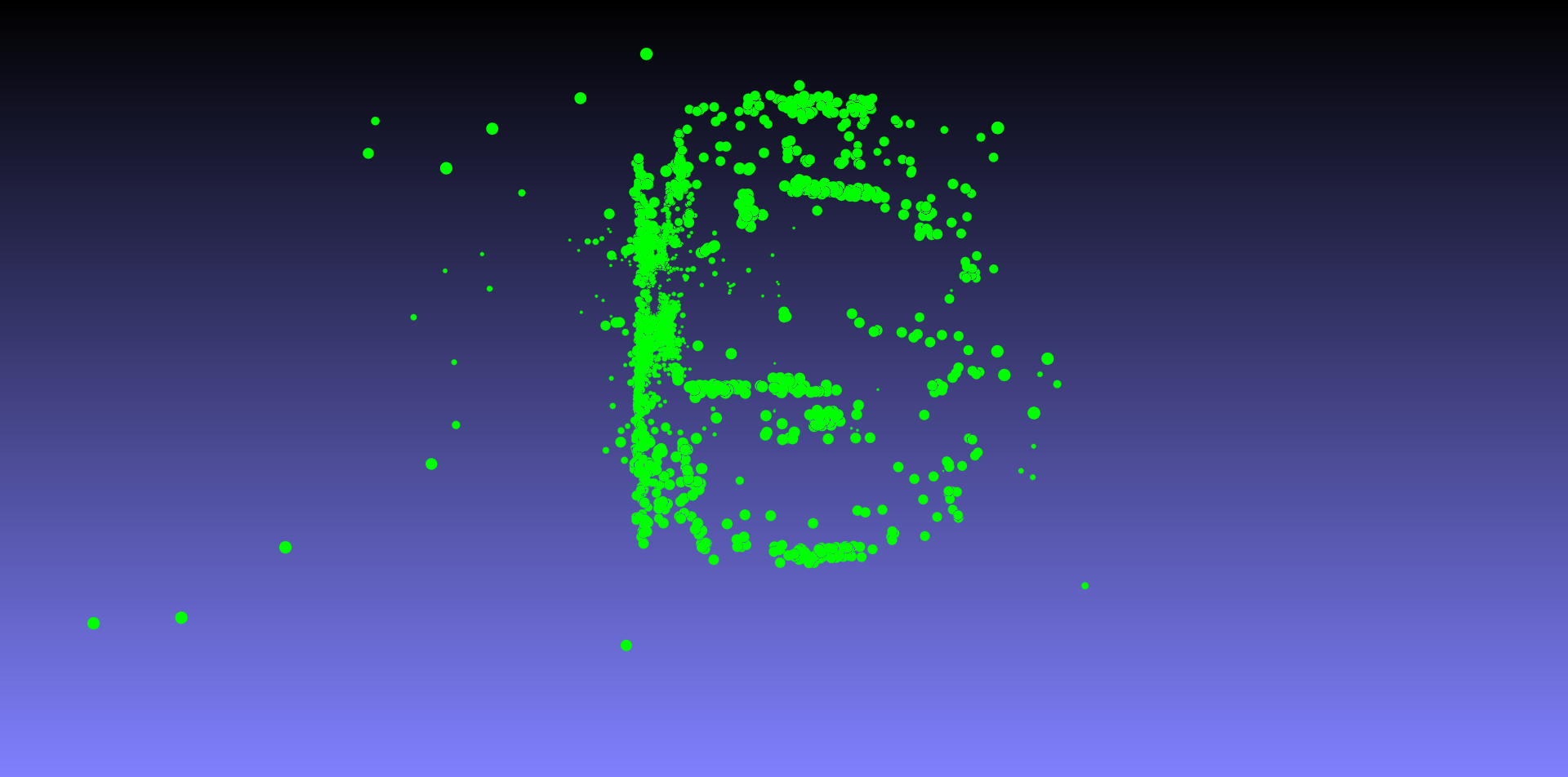}}
	\caption{Image sequence and views of the reconstructed point cloud. (a)-(j) images from the recording of one vehicle, (k) view in similar angle as the recording setting, (l) side view of the reconstruction, (m) top view of the reconstruction, (n) front view of the reconstruction. (k-n) are visualized with Meshlab~\cite{meshlab}.}
	\label{fig:reconstruction}
\end{figure*}
\subsection{2D CNN Classifiers}
Since the seminal work of~\cite{rel-alexnet}, Convolutional Neural Networks (CNN) have become the standard tool for image classification due to their impressive performance. While researchers have been working on modeling the visual cortex with convolutional networks for some time before~\cite{rel-neocognitron,rel-mnist}, the computing capabilities to train networks with millions of parameters have become available only in recent years. In the last years significant improvements have been made due to e.g. ReLU~\cite{rel-relu}, dropout~\cite{rel-dropout}, batch-norm\cite{rel-batchnorm}, residual connections~\cite{rel-resnet}, and so on. In our method we use a popular network called VGG-16 that was proposed in~\cite{rel-vgg16} due to its widely acknowledged representation capabilities. Pretrained weights are available online so we used it as starting point for our experiments.
\subsection{3D CNN Classifiers}
More recently 3D CNN classifiers were proposed that operate on volumetric data instead of image data. These general deep networks are designed to work on arbitrary types of objects~\cite{rel-voxnet,rel-vol3d}. However, they are limited in model size and complexity. Other works propose to change the underlying data structure, e.g.~\cite{rel-octnet} use octrees to reduce the amount of memory and computing power needed for every convolutional layer. However, their maximum network input size is $64^3$ mainly due to memory limitations. This input size is not suitable for our sparse representations, where we want to focus on fine-grained differences between vehicle reconstructions. There are also works which operate directly on point clouds instead of volumetric renderings~\cite{rel-pointnet,rel-pointnet2}. These works rely on dense point clouds while in our case only sparse 2.5D information is available.\\
A different approach is taken in~\cite{rel-shape}, where the authors train an ensemble of CNNs, where each CNN learns a view specific classifier rendered from poses surrounding a 3D shape model. While the idea of projecting 3D information is somehow related to our approach, in practical applications it is often not possible to render multiple views of the same object. In our setting we are additionally limited to a single viewing angle.\\
3D CNN classifiers can also be found in the area of medical imaging, where several 2D recordings are usually registered and stored as a 3D volume. These CNNs solve specific tasks related to certain organs or diseases and are strictly limited to this use case~\cite{rel-medical3d}.
\section{3D RECONSTRUCTION}
\begin{figure*}[tbhp]
	\centering
	\subfloat[image]{%
		\includegraphics[width=0.24\linewidth]{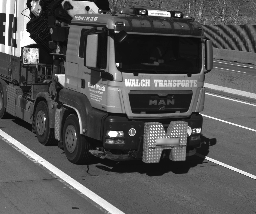}}
	\label{1a}
	\hfill
	\subfloat[reconstructed points]{%
		\includegraphics[width=0.24\linewidth]{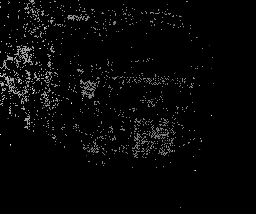}}
	\label{1b}
	\hfill
	\subfloat[reconstructed lines]{%
		\includegraphics[width=0.24\linewidth]{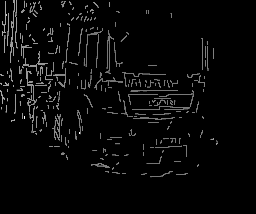}}
	\label{1c}
	\hfill
	\subfloat[reconstructed points+lines]{%
		\includegraphics[width=0.24\linewidth]{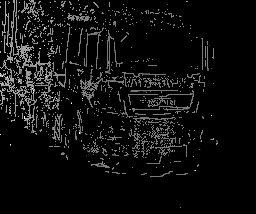}}
	\label{1d}
	\caption{Exemplary result of the reprojection. We use the camera center to reproject the reconstructed 3D points and lines into the 2D camera view. For every input image (a) three sparse reprojections are possible: (b) reprojected points, (c) reprojected lines and (d) combination of both. The depth is encoded by distance to the first camera center, where brighter colors denote a greater distance. Lines are well suited to capture vehicle structures and points allow for a slightly denser reconstruction. Combination of both yields the biggest accuracy improvement, as shown in our experimental section (Sec.\ref{sec:experiments}).}
	\label{fig:reprojection}
\end{figure*}
The aim of 3D reconstructions is to recover the three-dimensional structure of an object or scene from 2D images captured by one or more cameras. One prominent method is SfM, where camera poses are estimated and 3D points are triangulated from multiple camera views. We apply SfM to reconstruct vehicles for the task of vehicle classification on highways with a mobile vision system that is equipped with a single camera. This camera is mounted in a certain angle to the roadway and records the passing vehicles. In such a typical highway scene, as shown in Fig.~\ref{fig:reconstruction}, only parts of the street and of the vehicle are visible within one image frame. Therefore we integrate the 3D structure obtained by reconstruction into the classifier to capture vehicle properties that are not available in single images. For an optimal setting, we first calibrate the camera (Sec.~\ref{rec-calibration}). In contrast to standard SfM approaches, where the camera moves within the scene, our camera remains fixed and the vehicles we want to reconstruct pass the camera. For the reconstruction process this does not really make any difference, as the algorithm assumes the object to be static and the camera position is calculated relative to the object position. Every picture taken by the camera looks like a new camera to the SfM pipeline. As the vehicle passes the camera in constant direction, this generates a camera trajectory where the displacement between the virtual cameras is constant. Due to the high velocity of the passing vehicles, the number of frames available for reconstruction is very limited (typically less than or equal to $10$). This poses some challenges especially for the matching process, as many matches are found on the background. We exploit our knowledge of the scene to reduce the amount of background matches (Sec.~\ref{rec-context}). After reconstructing all of the recorded vehicles (Sec.~\ref{rec-rec}), we transform the point clouds into a common coordinate system by scaling it with a known scene element (Sec.~\ref{rec-mapping}).
\subsection{Calibration}
\label{rec-calibration}
Most 3D reconstruction algorithms require a calibrated camera system, especially when multiple cameras are used. As we reconstruct the 3D models of the passing vehicles from only one camera, the calibration is limited to the lens of this camera (intrinsic parameters). We use the toolbox provided by the authors of~\cite{calibration-toolbox} to calibrate the intrinsics. While it could be beneficial for specific tasks like metric vehicle volume estimation, we refrain from a metric calibration with the real world scene. Thus, our system can be easily set up at different positions without cumbersome manual calibration from the human operator. However, with known metric size of any scene part, this calibration could be easily added at anytime afterwards. For example, instead of taking a section of the middle line as reference like in Sec.~\ref{rec-mapping}, one could place a marker stick of certain length on the road and take one picture. With this known length, the reconstruction could be mapped into a metric coordinate system.
\subsection{Exploiting Context}
\label{rec-context}
While the practical setting we face in this work comes with some disadvantages (few images, fixed view), we exploit knowledge about two scene properties to improve the reconstruction results: Static camera and driving direction.
\\\textbf{Static camera.} As the camera is static, we remove false matches between two images by setting a threshold $d_p$ that determines the minimum distance in pixels a matching keypoint must have moved between two frames. A value below this threshold leads to removal of this match. In our experiments, we set this threshold to $50\text{px}$.
\\\textbf{Automatic estimation of driving direction.} Vehicles pass the camera driving in a certain direction. Thus, the movement of correctly matched keypoints must also correlate with this direction.
To automatically determine the moving direction, we extract lines from the captured scene images and apply Hough transform to discover the most prominent angle that corresponds to the main driving direction. We allow some deviation to make sure no correct matches are excluded. In our experiments the valid angles range from $320^\circ$ to $20^\circ$, where $0^\circ$ is the horizontal line.
\subsection{Reconstructing Points and Lines}
\label{rec-rec}
We use three types of 3D input data in our experiments. The first is a 3D point cloud and results from a standard SfM pipeline. As we deal with vehicles, the objects we want to classify are rigid and of cuboid shape. It seems natural to use 3D lines to describe the vehicles. We use~\cite{rel-l3d++}, which takes the camera poses from the SfM and generates a 3D line model. We use this as second type of input. As a third input type, we merge lines and points into one model, which is straightforward as both lie within the same coordinate system. While the points tend to capture a denser model of the vehicle, the lines are better representing the vehicle structure. In our experiments, we found that a combination of both yields the best results. See Fig.~\ref{fig:reprojection} for comparison.
\subsection{Aligning the Reconstructions within one World Coordinate System}
\label{rec-mapping}
\begin{figure} 
	\centering
	\subfloat[Schematic overview]{%
		\includegraphics[width=0.59\linewidth]{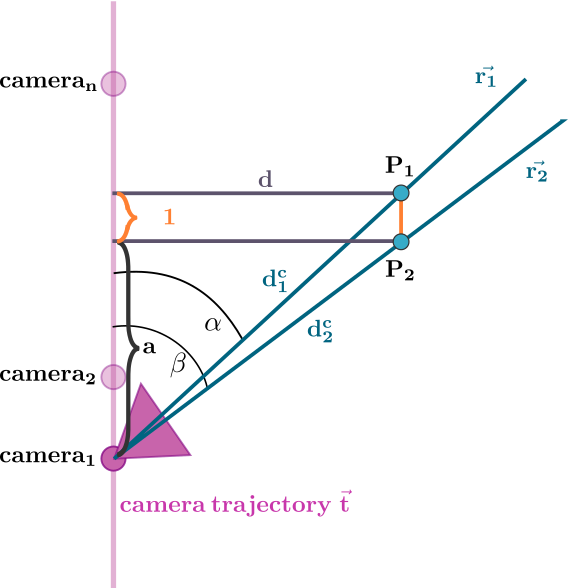}}
	\hfill
	\subfloat[Line in 3D]{%
		\includegraphics[width=0.40\linewidth]{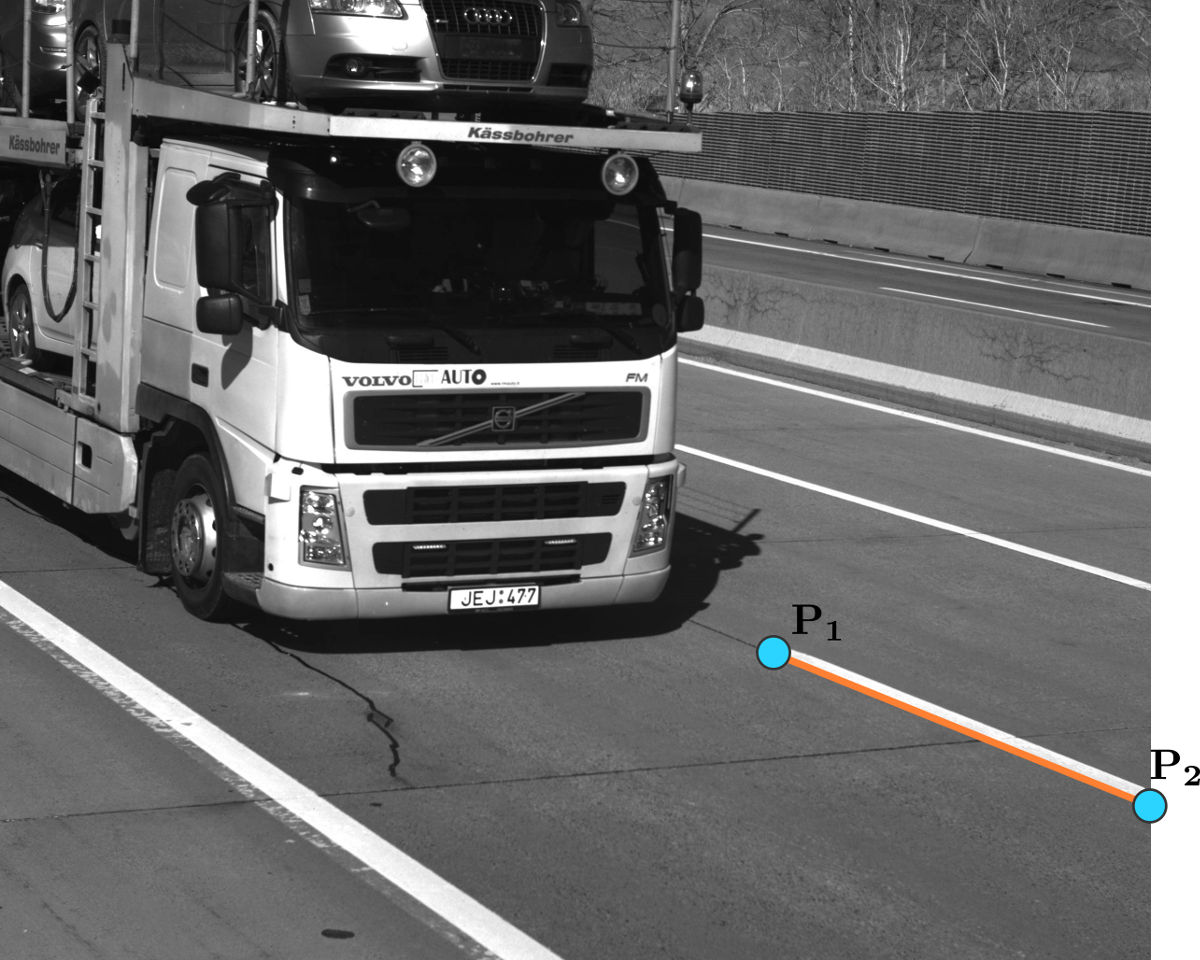}}
	\caption{Aligning the reconstructions. We exploit knowledge about the scene and select a line that is parallel to the driving direction of the vehicles and thus parallel to the camera trajectory of the reconstruction. With this reference line we can scale all models within the same world coordinate system, preserving the proportions and keeping depth reprojections consistent.}
	\label{fig:trigonometry}
\end{figure}

The SfM outputs camera poses that are equidistantly distributed along a trajectory. As the vehicles pass the camera with different speeds, the scale of the reconstructions differs and the depth information is not directly comparable. To resolve this issue, the reconstructed vehicles are mapped into a common world coordinate system. We translate the model to move the first view of the camera trajectory to the origin $(0,0,0)$ and choose a line on the street that is parallel to the driving direction and consequently also parallel to the camera trajectory in 3D space. We set the line length to $1$ in our 3D world coordinate system and require the camera distance to this line to be the same for all reconstructions. We then recover the 3D position of the points and use them to calculate the scale of the current reconstruction. With this scale we can transform the 3D model such that all models are equally scaled and made comparable.\\
To be more specific, the points $P_1$ and $P_2$ spanning the line are visible in the first camera. $P_1$ and $P_2$ lie on the rays $\vec r_1$ and $\vec r_2$ casted from the camera center at distances $d_1^c$ and $d_2^c$. The camera trajectory direction $\vec t$ is parallel to the 3D line at a distance of $d$. The distance from the camera center to the projection of $P_2$ onto the camera trajectory is denoted with $a$. Consequently, the distance from the projection of $P_1$ onto the trajectory is $a+1$, as we set the line length to $1$. The angle $\alpha$ describes the angle between the camera trajectory and the ray $\vec r_1$ from camera center to point $P_1$ and $\beta$ corresponds to the angle between camera trajectory and ray $\vec r_2$ to point $P_2$.With these preliminaries we can recover the scale by deploying trigonometric functions. To get the angles $\alpha$ and $\beta$, we first resolve the following equations:
\begin{align}
\alpha = \arccos \frac{\vec t \cdot \vec r_1}{|\vec t| |\vec r_1|}\:\:,\:\:\beta = \arccos \frac{\vec t \cdot \vec r_2}{|\vec t| |\vec r_2|}
\label{eqn:trigonometry1}
\end{align}
We can then use the angles to calculate $a$ by solving the set of equations
\begin{align}
\begin{array}{l}
\tan \beta = \frac{d}{a} \\
\tan \alpha = \frac{d}{a+1}
\end{array}
\label{eqn:trigonometry2}
\end{align}
for a. This results in
\begin{align}
a = \frac{1}{\frac{\tan \alpha}{\tan \beta}-1}.
\label{eqn:trigonometry3}
\end{align}
Now we can calculate the distance $d$ for one of the points by inserting into either 
\begin{align}
d = \tan \beta \cdot a\:\:\:\:\text{or}\:\:\:\:d = \tan \alpha \cdot (a+1).
\label{eqn:trigonometry4}
\end{align}
We then calculate the distances from the points to the camera center with 
\begin{align}
\begin{array}{l}
d_1^c=\sqrt{d^2+(a+1)^2}\\
d_2^c=\sqrt{d^2+a^2}.
\end{array}
\label{eqn:trigonometry5}
\end{align}
Finally, the scale is defined by 
\begin{align}
s=1/d_1^c.
\label{eqn:trigonometry6}
\end{align}
This way we use the distance $d_1^c$ from camera to $P_1$ as reference length for scaling. Fig.~\ref{fig:trigonometry} visualizes the procedure.

\section{VEHICLE CLASSIFICATION WITH DEPTH PRIOR}
Our classification method is based on the combination of 2D and 3D information. We reflect this in the design of our CNN that we employ for classification. The input images have two channels, of which the first contains the grayscale image captured by the camera and the second contains a depth reprojection from the 3D point cloud. We alter the CNN structure to incorporate an auxiliary branch that helps to classify the vehicles.
\subsection{Depth Reprojection}
One CNN input is a reprojection map including points and/or lines projected from the 3D model into the 2D camera view. Any reprojected pixel value represents the depth measured from the origin of the world coordinate system, in our case the first camera. A projection example can be seen in Fig.~\ref{fig:reprojection}. To capture as much vehicle structure as possible, we set the parameters in a non-strict fashion to allow an imperfect reconstruction from the limited number of available images. This results in some errors that cause entirely wrong reprojected depth values. At first, this poses a problem to the optimization task, where we optimize on a very sparse number of values, however our network is able to deal with this through a selected loss function, as described in the next section.
\subsection{CNN Model Structure}
As basis network structure, we use a VGG-16 model~\cite{rel-vgg16} pretrained on ImageNet\cite{imagenet} and finetune it on our data. As the pretrained model is for RGB images, we replicate the grayscale image. We modify the network structure and add an auxiliary branch, which tries to estimate the depth from the 2D input image. We use the activation maps after the \textit{conv5} layer as inputs to this branch. We then add a $1\times1$ convolution to reduce the number of activation maps from $512$ to $1$. On top of this layer, we add a deconvolution layer that upscales the activations to the input size. We then calculate an auxiliary loss between these upscaled activations and the input reprojection image. We also set the number of outputs of the softmax layer to $6$, according to the number of vehicle classes. Fig.~\ref{fig:vgg-model} shows the changes made to the original network structure.
\\\textbf{Auxiliary Loss.} Due to the imperfect reconstruction, there exists a limited number of wrong depth values within the reprojection map. These wrong values could heavily impact the optimization during training if using a standard $\mathcal{L}_2$ loss. To avoid instability problems during optimization, we use a Huber loss to compensate for errors in the reprojection map. The Huber loss has a linear part for absolute values larger than $\delta$ and is defined as 
\begin{align}
\ell_H(x) = \left\{
\begin{array}{cl}
\frac{1}{2} x^2 & \text{for }|x| \le \delta, \\
\delta \left(|x|-\frac{1}{2}\delta \right) & \text{otherwise.}
\end{array}
\right.
\label{eqn:huber_loss}
\end{align}
In our case, $\delta$ is the difference of the depth reprojection and the CNN depth prediction. The reprojection map is very sparse, therefore we mask the loss only for pixels with known depth. In our experiments, we set $\delta=0.1$. We weight the auxiliary loss $\ell_{\mathbf{aux}}$ with a parameter $\lambda_{\mathbf{aux}}$ and add it to the classification loss $\ell_c$, to train our network with the loss
\begin{align}
\ell = \ell_c + \lambda_{\mathbf{aux}} \cdot \ell_{\mathbf{aux}}.
\label{eqn:training_loss}
\end{align}

\section{EXPERIMENTS}
\label{sec:experiments}
To show the efficacy of our method, we evaluate multiple experiments on a dataset of over $400$ vehicles. To set a baseline, we deactivate the auxiliary branch and train the network without depth information. We compare the baseline to results with our three input variations (points, lines, both) and report the classification accuracy on a test set for single images and sequence-wise. We train our network on a NVIDIA TITAN Xp GPU with a batch size of $b=64$ until convergence and employ a stochastic gradient descent (SGD) optimizer with a learning rate of $1e^{-3}$ and momentum set to $0.9$.
\\\textbf{Dataset.} 
Our dataset consists of $439$ sequences comprising $2688$ images. The $439$ vehicles are labeled into $6$ different categories: \textit{special transport} ($122$ images), \textit{car} ($448$), \textit{camper} ($9$), \textit{van} ($222$), \textit{truck} ($571$) and \textit{semitrailer} ($1316$). For our experiments, we split the dataset $80\%$ for training and $20\%$ for testing.
\\\textbf{Results}. 
Table~\ref{tab:results} shows the experimental results of our method for all three input variants (points, lines and both). We report the accuracy on the test set image-wise and sequence-wise. For the latter case, we first classify all images of a sequence (typically 3 to 10) and count it as correct if more than half of the images are correctly classified. Our method improves over the baseline without auxiliary branch for all input types. The combination of both, points and lines, yields the highest accuracy.
\begin{table}[htbp]
	\caption{Experimental Results: Results obtained for all input variants with $b=64$, $\lambda_{\mathbf{aux}}=1e^{-2}$, $\delta=0.1$}
	\begin{center}
		\begin{tabular}{|c|c|c|}
			\hline
			\textbf{Input variant} & \textbf{Image Accuracy} & \textbf{Sequence Accuracy}\\
			\hline
			baseline & $87.90\%$ & $91.01\%$\\
			\hline
			lines & $89.01\%$ & $93.26\%$\\
			points & $89.39\%$ & $93.26\%$\\
			points+lines & $\mathbf{90.13\%}$ & $\mathbf{94.38\%}$\\
			\hline
		\end{tabular}
		\label{tab:results}
	\end{center}
\end{table}
\section{CONCLUSION}
In this paper, we present a method that exploits 3D information to improve 2D classification accuracy. We reconstruct 3D models of vehicles passing a static camera and encode the depth in a 2D reprojection. These reprojections are used in an auxiliary branch of our CNN, where the network reconstructs depth values and acts as a regularizer. We show that our method improves classification accuracy for all three input variants (points, lines, both) over the baseline without 3D information on a real world dataset. At test time, our method does not need 3D information and can thus be employed on mobile vision systems.

\end{document}